\RequirePackage{fix-cm}
\documentclass[smallcondensed]{svjour3}     
\smartqed  
\usepackage{graphicx}
\usepackage{subcaption}
\usepackage{xcolor}
\usepackage{amsfonts}
\usepackage{amstext}
\usepackage{amsmath}
\usepackage{mathtools}
\usepackage{calc}
\usepackage{bigints}

\usepackage[pagebackref,colorlinks=true,breaklinks=true,urlcolor= magenta,bookmarksopen=false]{hyperref}
\hypersetup{pdftitle={Title},pdfauthor={Author},linkcolor=blue,citecolor=magenta}

\usepackage{todonotes}

\newcommand{\etalcite}[1]{{\emph{et al.}$\!$}~\cite{#1}}

\journalname{Journal of Intelligent \& Robotic Systems}
\begin{document}

\title{User-oriented Natural Human-Robot Control with Thin-Plate Splines and LRCN\thanks{FAPEAL/CAPES grant 05/2018}
}


\author
{
    Bruno Lima \and Lucas Amaral \and Givanildo Nascimento-Jr \and  Victor Mafra \and Bruno Georgevich Ferreira \and Tiago Vieira \and Thales Vieira
}

\authorrunning{Lima et al.}


\institute{Bruno Lima \and Lucas Amaral \and Givanildo Nascimento-Jr \and  Victor Mafra \and Bruno Georgevich Ferreira \and Tiago Vieira \and Thales Vieira \at
Institute of Computing, Federal University of Alagoas \\
\email{\{bgcl, lafa, glnjunior, vmhf, bgf, tvieira, thales\}@ic.ufal.br}
}

\date{Received: date / Accepted: date}

\maketitle

\begin{abstract}
We propose a real-time vision-based teleoperation approach for robotic arms that employs a single depth-based camera, exempting the user from the need for any wearable devices. By employing a natural user interface, this novel approach leverages the conventional fine-tuning control, turning it into a direct body pose capture process.
The proposed approach is comprised of two main parts. The first is a nonlinear customizable pose mapping based on Thin-Plate Splines (TPS), to directly transfer human body motion to robotic arm motion in a nonlinear fashion, thus allowing matching dissimilar bodies with different workspace shapes and kinematic constraints. The second is a Deep Neural Network hand-state classifier based on Long-term Recurrent Convolutional Networks (LRCN) that exploits the temporal coherence of the acquired depth data.
We validate, evaluate and compare our approach through both classical cross-validation experiments of the proposed hand state classifier; and user studies over a set of practical experiments involving variants of pick-and-place and manufacturing tasks. 
Results revealed that LRCN networks outperform single image Convolutional Neural Networks; and that users' learning curves were steep, thus allowing the successful completion of the proposed tasks. When compared to a previous approach, the TPS approach revealed no increase in task complexity and similar times of completion, while providing more precise operation in regions closer to workspace boundaries.
\end{abstract}

\section*{Declarations}
\subsection*{\textbf{Funding}}
This research was supported by FAPEAL/CAPES grant 05/2018.

\subsection*{\textbf{Conflicts of interest/Competing interests}}
The authors have no conflicts of interest to declare that are relevant to the content of this article.

\subsection*{\textbf{Availability of data and material}}
Our collected depth images dataset is publicly available through a link provided in Section \ref{sec:results_classifier}.

\subsection*{\textbf{Code availability}}
Code used in this article is stored in GitHub and will be made publicly available.

\subsection*{\textbf{Authors' contributions}}

\textbf{Bruno Lima}: Methodology, Software, Investigation, Writing - original draft preparation, Writing - review and editing; \textbf{Lucas Amaral}: Methodology, Software, Investigation, Writing - review and editing; \textbf{Givanildo Nascimento-Jr}: Methodology, Software, Investigation, Writing - original draft preparation, Writing - review and editing; \textbf{Victor Mafra}: Methodology, Investigation, Writing - original draft preparation, Writing - review and editing; \textbf{Bruno Georgevich Ferreira}: Methodology, Investigation, Writing - review and editing; \textbf{Tiago Vieira}: Conceptualization, Investigation, Writing - review and editing, Resources, Supervision; \textbf{Thales Vieira}: Conceptualization, Investigation, Writing - review and editing, Resources, Supervision.

\subsection*{\textbf{Ethics approval}}
Ethics approval was not needed since participants were not explicitly identified, nor asked to perform dangerous, harmful or invasive tasks.

\subsection*{\textbf{Consent to participate}}
Informed consent was obtained from all individual participants included in the study for research purposes.

\subsection*{\textbf{Consent to publish}}
Informed consent was obtained from all individual participants regarding publishing their data and photographs for research purposes.

\clearpage

\section{Introduction}
\label{sec:intro}


Human-Robot Interaction (HRI) comprises the field that allows humans and robots to communicate with the aim of accomplishing well-defined goals. This interaction can be either close or distant, depending on the aspects of space and time involved. According to~\cite{goodrich2008human}, two paradigms of human-robot interaction were always present within the robotics context: (1) teleoperation and (2) supervisory control. Teleoperation is the act of remotely operating a robot, either mobile or fixed, without necessarily being physically present in the same environment. Supervisory control is the process of supervising (watching and possibly acting for correction) the behaviour of an autonomous system or process. 

A common application example for discussed fields takes place in manufacturing industries with human cognitive inspection~\cite{villani2018survey}. A surveillance mobile robot equipped with a camera and a robotic arm may inspect and take simple actions such as activating an emergency alarm or turning up a crank. Another application would be aiding search and rescue tasks in potentially harmful places~\cite{delmerico2019current}. A mobile robot may survey an unknown area while looking for victims and assessing structural stability. In both these applications, advantages reside in the ability to address complex problems from unstructured and possibly unknown environments, combining human cognition with sensory tools and low-level autonomous behavior of robots.

Teleoperation is also one subset of Learning from Demonstration (LfD), besides kinesthetic teaching and passive observation. According to Ravichandar~\etalcite{doi:10.1146/annurev-control-100819-063206}, LfD is the process of a robot acquiring new skills through imitating an expert. It is appropriate for situations where ideal behavior might not be easily scripted or defined as an easy optimization problem. By trying to replicate the expert motions, robots might implicitly learn task constraints. When combined with exploration-based methods and exhaustive simulation, adaptive behavior for unstructured environments can be achieved. Furthermore, in cases where imitator and expert have dissimilar bodies, a kinetic mapping between motions becomes necessary, commonly referred to as motion \textit{re-targeting}. A similarity metric can be used to evaluate the quality of re-targeting, which can be performed either in joint or Cartesian spaces.

As we face the increasing presence of robots within the human workspace, a concern that naturally emerges is the guarantee of safety for human individuals. As stated in~\cite{zacharaki2020safety}, it is necessary to address physical, psychological, and social influence over people who get exposed to robots' perception, cognition, and action. It is also shown that there was a boom in RGB-D (RGB along with a depth channel) sensory devices development starting from 2010, causing a noticeable interest in perception-enabled safety systems. As input data, these systems would use accurate real-time depth information from the observed workspace, which may include the environment, humans, and other robots. 
When compared to common RGB cameras, the main advantage of RGB-D sensors is a higher invariance to color and lighting conditions. One direct implication is to use depth data for human skeleton cognition and pose inference with Deep Neural Networks. Vision-based human pose detection allows for high-level interactions, such as controlling the position of a robot based on human body movement. The main advantage resides in direct motion control, exempting the user from additional hardware components such as complex joysticks or wearable devices. Concerning teleoperation, using vision-based control naturally brings into light the re-targeting problem: how to effectively map human motion to robot movement.

In the discussed scenario of Human-Robot Interaction (HRI), we aim to address the problem of finding a good vision-based position control through a natural user interface that exploits human motion for teleoperating a robotic arm, using only non-intrusive RGB-D data from the user as sensory input, with no additional controllers or wearable devices. As a study case, we implement a complete position control system for a robotic arm and its end-effector using Robot Operating System (ROS) as a communication backend.

Our main contributions are the following:
\begin{enumerate}
    \item A human hand state classifier based on  Long-term Recurrent Convolutional Networks (LRCN) that exploits the temporal coherence of the problem, thus allowing a reliable robot gripper state control;
    \item A novel pose mapping approach based on Thin-Plate Splines (TPS) to transfer human body poses to the robotic arm poses. We highlight that the proposed pose mapping can be individually trained to fit each particular user body. 
\end{enumerate}

As corroborated in our experiments, the combination of these parts provides a natural user interface that showed to be an intuitive and responsive alternative for teleoperation, with enhanced precision in critical workspace regions, such as their boundaries.

\section{Related work}
\label{sec:related-work}

The problem of robotic control in harmful environments has led to the development of different approaches over the past decades. In~\cite{9198439}, this problem is addressed by using electromyogram (EMG) signals recorded from the muscles of the elbow. The proposed applications were assisting elderly people, supporting people with physical disabilities, or teleoperation in hazardous sites. They applied the methodology in a 5 degrees of freedom (DOF) anthropomorphic robotic manipulator, by mimicking the exact joint angles from the user's arm. To accomplish it, they used two accelerometer sensors coupled to the user's arm alongside the elbow EMG readings as input data. Different from us, they addressed the problem by replicating the full user's arm joint values in a low-cost robotic arm system, besides using EMG and acceleration sensors. In our case, we simplified the problem to address only the gripper position mapping, using a fixed down-side orientation which is proper for most pick-and-place tasks.

In~\cite{10.1145/3173386.3177084}, the problem is addressed from a teleoperated robotic nurse point of view. Their focus was on human-robot collaboration in medical environments, making a study about the acceptability of a teleassistive robotic system among healthcare professionals. Their methodology consisted of testing 3 different control modes for a manipulative goal: haptic control, gesture control, and voice commands. Their gesture control method used a Kinect to capture the user's upper body skeleton and pose information, alongside two EMG readers (in armband form) to infer the user's hand state (open or closed). Armbands also provided gyroscope information to detect the roll/pitch motion of the user's hand to control the robot's mobile base. They performed the validation experiment with 11 pre-med and bio-engineering students as participants, evaluating two main parameters in the form of a Likert scale survey: (1) difficulty in control and (2) safety. Different from their approach, in our experiments the robot's base was fixed. Moreover, we did not use any wearable device for the robot manipulator control. Both the user pose and the hand state were inferred directly from the RGB-D data. On the other hand, we also used Likert scales for our participants' survey, but with different measured aspects (smoothness of movement and difficulty near workspace boundaries).

The problem of kinematic re-targeting is discussed in~\cite{chen2020virtual}, while proposing a virtual-joint based similarity criteria for evaluation mapping performance. Two platforms were implemented to experimentally verify the flexibility and the effectiveness of the proposed metric, both using the MATLAB framework. Specifically, they used the Robotic Toolbox for model visualization and the Optimization Toolbox for kinematics mapping. Two robots with 6 DOF were used. 
Similar to our project, a transformation mapping from the user's left arm to the robot was the subject of the experiments in three steps: user pose capture, kinematics transformation and, lastly, transfer to the real robot. Differently from them, our work used a metric-less non-linear mapping by directly specifying points from each Cartesian workspace (demonstrator and imitator). They also did not take into consideration the hand state mapping for the robot gripper.

In~\cite{lipton2017baxter}, the theme of teleoperation is addressed while applied to a manufacturing environment aided by virtual reality (VR) techniques. Using the concept of a Virtual Reality Control Room (VRCR), the user's state is mapped to the VRCR, and the VRCR is mapped to the robotic system, decoupling the sensory stimuli (which is given from the virtual environment) from the robot communication. They tested the system using the Oculus Rift with Touch controllers, along with a Baxter robot. Baxter provided camera feedback for the user in both the stereo-camera placed at the head of the robot and an end-effector camera for each gripper. They assessed the effect of a high-latency network, through the simulation of a limited system connectivity. Similarly to us, they also tested the system for pick-and-place and assembly tasks, measuring their time of completion. Their sub-tasks (which one could associate to the atomic operations in our work) involved the interaction with two Lego cubes, which were: approaching, grasping, aligning, and assembling (stacking one on top of another). If the user made an unrecoverable error, miss-aligned the blocks, or took longer than 6 minutes for each sub-task, the attempt was recorded as a failure. Different from us, they assessed the mapping for both human arms, although our system expands naturally for both arms. Also, we did not need any wearable devices like Oculus Rift or Touch Sensors, relying only on the Depth Sensor and visual feedback (i.e.: a common RGB camera).

As proposed in~\cite{visapp19}, one way of addressing the problem of teleoperation is by using a depth sensor to track and mimic the teleoperator movements, leading to a natural movement interface between user and robot. In that work, authors proposed to exploit skeletons extracted from a depth sensor to map the user's hand position (w.r.t. the shoulder frame) to the robot gripper position (w.r.t. the base frame), providing a mapping that is invariant to user location and orientation in the camera space. However, the proposed mapping between a human arm pose space to a robotic arm pose space was a naive linear mapping, which is not efficient since both demonstrator and imitator workspaces have different volumes (former is a semi-sphere while the latter is a semi-cylinder, approximately) and thus different kinematic constraints. They also developed a hand-state classifier using depth images and Deep Neural Networks. As a depth sensor, they used a Kinect V1, along with a robotic arm (Denso VP6264) and a gripper (Robotiq 2F-85). 
Differently from that previous work, we propose a novel hand-state classifier based on the Long-term Recurrent Convolutional Networks (LRCN), and a novel pose mapping approach based on a Thin Plate Spline (TPS) formulation that learns from each user, through a training session, a customized non-linear pose map to appropriately transfer its body movements to the robotic arm movements.

\section{Overview}
\label{sec:overview}

\begin{figure*}[!ht]
\centering \includegraphics[width=\textwidth]{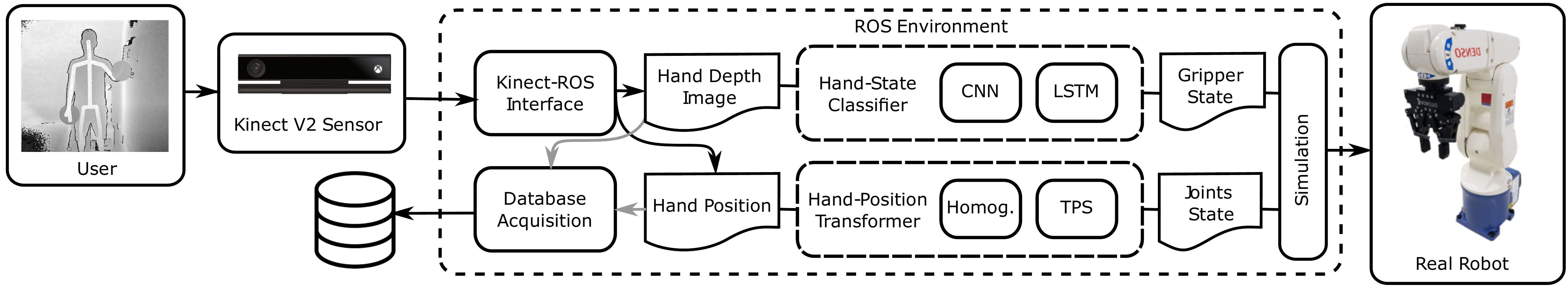}\\
\caption{Diagram of system architecture. Kinect One sensor is used as input sensor while robotic arm represents the output actuator. Processing steps run in a distributed manner, with nodes that communicate among them through a ROS Environment. Classification and pose transformation run independently from each other.}
\label{fig:diagram}
\end{figure*}

In this work, we focus on the problem of developing natural user interfaces to control a robotic arm using a depth sensor such as  Kinect V2. We face the challenge of developing an accurate and intuitive interface to control the position and the state of a robotic arm manipulator through human motion, without using any wearable device such as haptic gloves.
We take advantage of the similarity between the human superior limbs and the robotic arms, and propose to map the human hand position w.r.t. the shoulder to the robotic gripper position w.r.t its base; and the human hand state to the robot gripper state. With this goal, in addition to the robotic arm, the proposed method relies only on a Kinect V2 sensor, which could eventually be replaced by a similar depth sensor, and an average desktop computer.


The method is comprised of two main parts: a novel customizable motion mapping function based on Thin Plate Splines~\cite{bookstein1989principal}, which relies on a personalized user training session to fit each individual user body characteristics;
and a recurrent neural network trained to recognize whether the user hand is opened or closed from depth images. Fig.~\ref{fig:diagram} shows an overview of the method, which we detail in the following sections.

\section{\textbf{Natural Human-Robot Interface}}
\label{sec:basic_nhi}

The development of user interfaces for robotic arm teleoperation is challenging due to the high dimension of the robot pose space and the non-intuitive output of varying each 
joint angle. In this work we propose a novel method based on Thin Plate Splines (TPS, \cite{bookstein1989principal}) to smoothly and intuitively map human body poses to the robotic arm pose. We also rely on depth sensor skeletons to estimate human body poses but using a Kinect v2 in our experiments. 
The proposed mapping is customized for each user in an individual training session, to capture its specific anatomic limitations.

Following, we first describe the invariant human arm representation which was previously presented in~\cite{visapp19}. Then, we briefly describe the TPS formulation and how it is employed to map the human arm pose to the robotic arm pose. Finally, we detail a straightforward training process in which a user may customize the mapping to fit its body characteristics.

\subsection{Human Arm Representation} \label{sec:armrepresentation}

To provide invariance to translations and rotations of the body w.r.t. to the depth sensor, we employ the human arm pose representation presented in~\cite{visapp19}. Note that only the relative position of the hand w.r.t. the corresponding shoulder is exploited to control the robot.

\begin{figure}[!b]
\centering \includegraphics[width=\columnwidth]{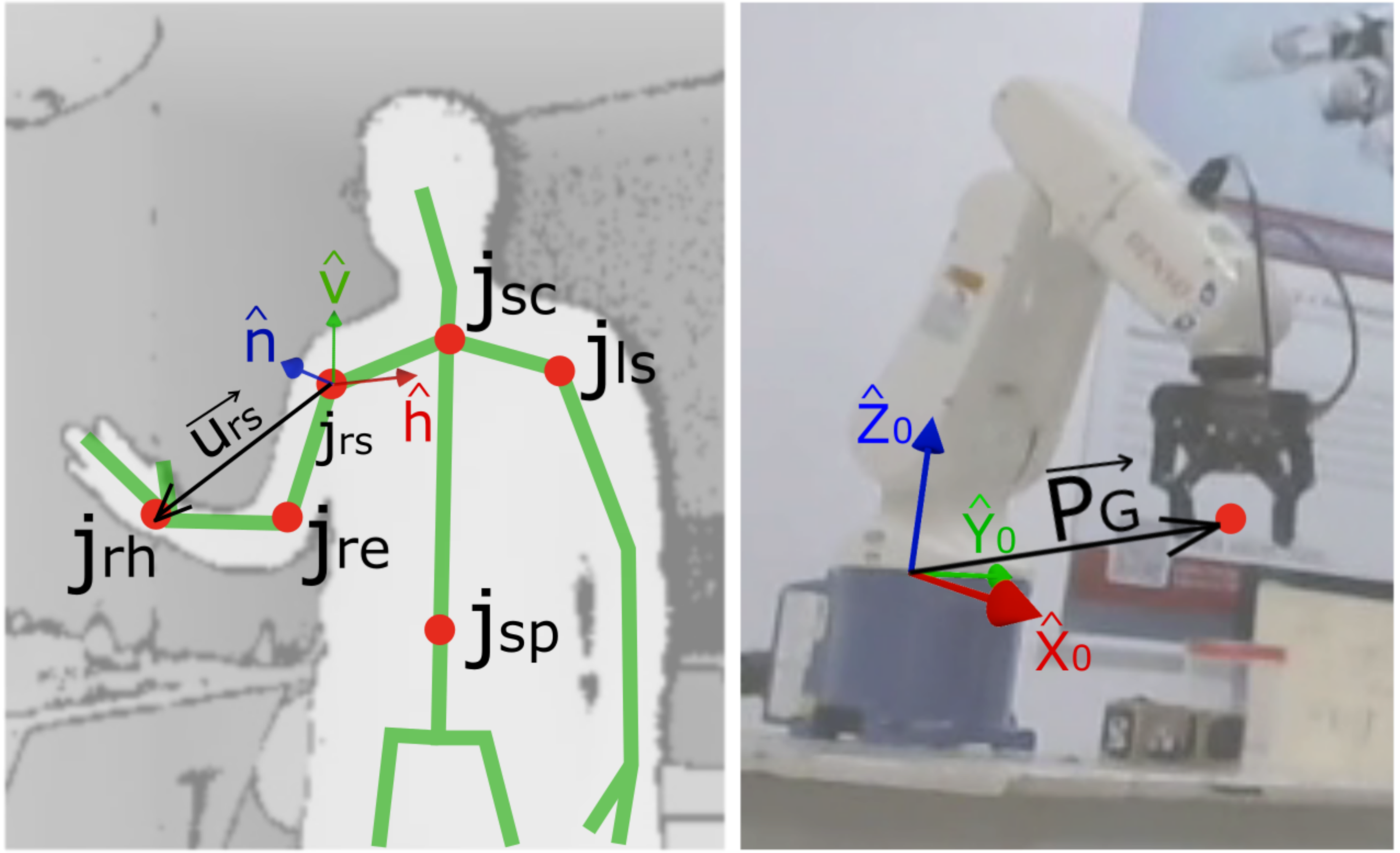}\\
\caption{Illustration of mapping between coordinate systems. Vector $u_{rs}$ from user's shoulder coordinate system (left image) is mapped to vector $P_G$ from robot base coordinate system (right image), located at gripper center.}
\label{fig:frames}
\end{figure}

First, we define a reference frame centered on the active shoulder, which we consider to be the right shoulder from now on. The following skeleton joints are used: right shoulder ($j_{rs}$); right elbow ($j_{re}$); right hand ($j_{rh}$); left shoulder ($j_{ls}$); shoulder center ($j_{sc}$); and spine center ($j_{sp}$). Used joints and vectors are illustrated in Fig.~\ref{fig:frames}, and discussed at following paragraphs.


The basis of the reference frame is given by three vectors related to the torso plane. Two of them represent the torso vertical and horizontal directions. More specifically, we compute vertical $\hat{v}$ and horizontal $\hat{h}$ unit vectors given by:
\begin{equation}
    \hat{v} = \frac{j_{sc} - j_{sp}}{\|j_{sc} - j_{sp}\|}, \quad \hat{h} = \frac{j_{ls} - j_{rs}}{\|j_{ls} - j_{rs}\|} \, .
\end{equation}
The last vector of the basis is the normal vector $\hat{n}$ of the torso plane, which can be easily computed by the normalized cross product:
\begin{equation}
\hat{n} = \frac{\hat{h} \times \hat{v}}{\| \hat{h} \times \hat{v} \|} \, .
\end{equation}

A right arm vector $\vec{u}$ is first computed within camera coordinate system, where:
\begin{equation}
    \vec{u}=\vec{j_{rh}} - \vec{j_{rs}}.
\end{equation}

Finally, the aforementioned invariance is achieved through a change of basis of $\vec{u}$ from the camera basis $\{\hat{c}_x, \hat{c}_y, \hat{c}_z\}$ to the reference frame basis $\{\hat{h}, \hat{v}, \hat{n}\}$ (see Fig.~\ref{fig:frames}):

\begin{equation}
\vec{u}_{rs} = \vec{u} \cdot
\left[
    \hat{h},
    \hat{v},
    \hat{n}
\right].
\label{eq:armlen_unormalized}
\end{equation}

%


\subsection{Human-Robot Pose Mapping}
\label{sec:ht}
In this subsection, we briefly describe the naive transformation proposed in~\cite{visapp19} to map the human arm pose to the robotic arm pose and discuss its limitations. Then, we present the novel mapping approach based on TPS.

\subsubsection{Naive affine pose mapping} \label{sec:naivemapping}
In this straightforward approach, the input is given by a normalized arm vector ($\vec{u}_{rs}'$), due to differences in users' arm lengths.
To perform this normalization, the sum of the lengths of the arm and forearm vectors is employed as follows (Fig.~\ref{fig:frames}):
\begin{equation}
\vec{u}_{rs}' = \dfrac{\vec{u}_{rs}}{\| j_{re} - j_{rs} \| + \| j_{rh} - j_{re} \|}  = [u_x', u_y', u_z'] \, .
\label{eq:armlen}
\end{equation}
Then, the following affine transformation is applied to mimic the human arm pose to the robotic arm pose:


\begin{equation}
    P_G = 
    \left [
    \begin{array}{c}
    \eta \omega_x u_z' + \delta_x \\
    \eta \omega_y u_x' + \delta_y \\
              \omega_z u_y' + \delta_z \\
    \end{array}
    \right ]
    \cdot
    \left [
    \begin{array}{c}
    \hat{X}_0\\
    \hat{Y}_0\\
    \hat{Z}_0
    \end{array}
    \right ]
\end{equation}
where:
\begin{itemize}
    \item $(\omega_x, \omega_y, \omega_z)$ is a constant mapping proportion vector;
    \item $(\delta_x, \delta_y, \delta_z)$ is a constant mapping displacement offset;
    \item $(\hat{X}_0, \hat{Y}_0, \hat{Z}_0)$ is the orthonormal basis of the robot coordinate system.
    \item $\eta \in \left\{-1, 1\right\}$ is a constant that represents a mirroring mode, allowing for mirroring horizontal and normal axes of user's body plane;
    \item $P_G$ is the position of the gripper w.r.t. the robot coordinate system, as shown in Figure~\ref{fig:frames}.
\end{itemize}
It is worth mentioning that we consider that the gripper orientation always stays pointing downwards since this is the default orientation for simple pick-and-place tasks.

Although fairly simple, this transformation requires a manual tuning of the aforementioned constants, which may vary depending on the user's preferences and robot motion boundaries. Besides, even after a careful calibration procedure, one of the following cases might stand: (1) the resulting calibration is restrictive and robot movement becomes limited to prevent mapping the robot gripper position outside its valid workspace; or (2) the calibration is more flexible, leading to robot gripper positions that must be projected to the workspace boundary. Both scenarios occur due to the inherently different geometries of the pose spaces (semisphere vs. semicylinder).

Also, different users might have substantially different body limitations concerning arm movement (body limits that may not be fixed by the arm length normalization discussed in Section \ref{sec:basic_nhi}). For instance, a specific user might not be able to reach his left side with his right arm as far as the others. Such limitations motivated the use of (1) a more flexible non-linear family of mapping functions; and (2) a customized training procedure to fit the mapping function to each specific user body.

\subsubsection{Thin Plate Splines Pose Mapping}
\label{sec:tps}

Thin plate Splines (TPS)~\cite{bookstein1989principal}, are a class of spline mapping functions that have been applied to distinct problems such as interpolation~\cite{hutchinson1995interpolating}, and non-rigid image~\cite{rohr1996point} and surface registration~\cite{brown2004non}.

Such mapping functions are easily computable, globally smooth, separable into affine and non-affine components, and provide the least possible non-affine warping component in the presence of mapping constraints.
Given two corresponding point sets $X = \{x_{1},...,x_{m}\}$ and $Y = \{y_{1},...,y_{m}\}$, Duchon~\cite{duchon1977splines} proved that there is a unique function $f \colon \mathbb{R}^n \mapsto \mathbb{R}$ that minimizes the bending energy given by 
\begin{equation}
\label{tps_1}
J = \bigintsss {\left(\sum_{i,j} f_{x_{i}x_{j}}^{2}\right)dx_{1} \dotsm dx_{n}} \, ,
\end{equation}
constrained to $f(x_i)=y_i^k , \,$ for $i=1 \dots  m$, where $y_i^k$ is the $k$-th coordinate of point $y_i$.

This unique function takes the form $xD + K(x)w$, where $x$ is a row vector containing the homogeneous coordinates of a point of $X$, $D$ is an affine transformation; $K(x)$ is an $m$-dimensional row vector in which $K_{i}(x)=U(\left | x - x_{i} \right |)$ is a Green's function; and $w$ is an $m$-dimensional column vector of non-affine warping parameters constrained to $M^tw = 0$. Here, $M$ is an $m \times n$ matrix where each row contains the coordinates of a point of $X$. 



Note that, given the corresponding point sets $X$ and $Y$, one has to find the TPS parameters $w$ and $D$. Fortunately, such a solution is given in closed-form as the solution of the linear system:
\begin{equation}
\label{eq:tps_3}
    \left( \frac{w}{D} \right) = 
    \left( 
        \begin{array}{c|c}
            \ K \    & \ M \ \\
            \hline
            \ M^t \  & \ 0 \ \\
        \end{array}
    \right)^{-1}
    \left( \frac{Y}{0} \right) \, ,
\end{equation}
where $K_{ij} = U(\left | x_i - x_j \right |)$. 


In the following, we describe how to train a TPS to map a user arm pose to the robotic arm pose, in a customized manner, to provide a smooth and adaptable Human-Robot Interface. 

\subsubsection{Customized TPS training} \label{sec:customtps}

As discussed in Section~\ref{sec:ht}, the naive affine transformation proposed in~\cite{visapp19} is unsuitable to map a semi-sphere to a semi-cylinder, besides having the necessity of fine-tuning parameters depending on user-specific body constraints.


We propose to compute a non-linear TPS pose mapping function based on constraints given by keypose correspondences between the user and the robot. Our hypothesis is that, by constraining a few relevant keyposes on the boundaries of the human and the robot working spaces, the other poses would be smoothly and intuitively mapped by the interpolation capabilities of the TPS.
More specifically, we propose a custom training of corresponding key poses for each specific user.

\begin{figure*}[ht]
\centering \includegraphics[width=\textwidth]{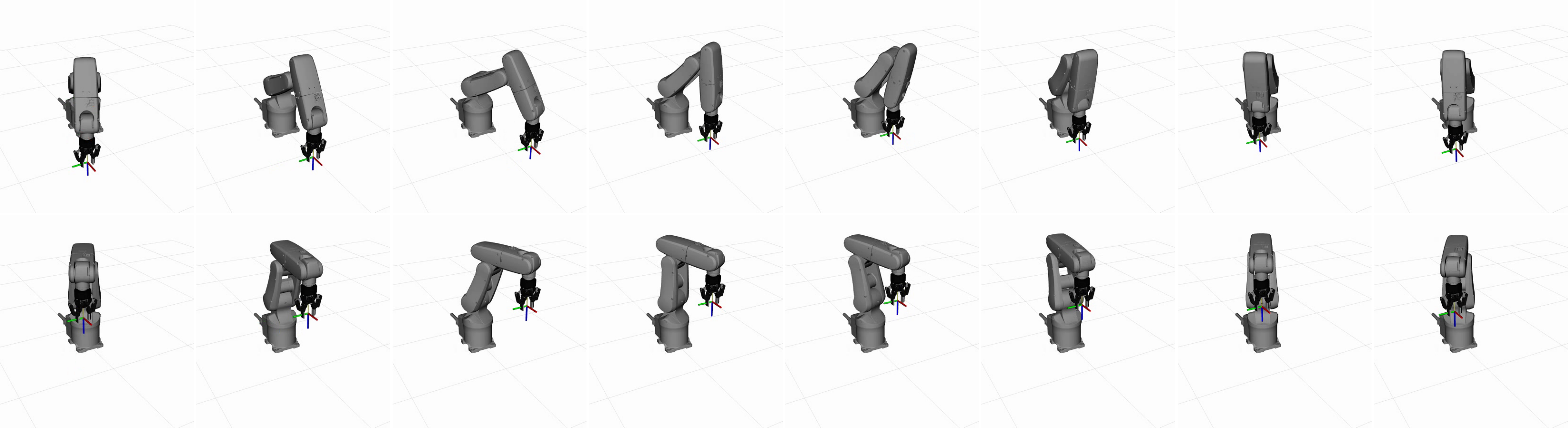}\\
\caption{Set of 16 robot poses used to train the TPS mapping model. Poses were organized into two different heights. Points were symmetrically chosen, at the boundary of the robot workspace.}
\label{fig:poses_robot_collage}
\end{figure*}

We empirically defined a set of 16 robotic arm poses that should be mapped to their respective human arm poses. To avoid extrapolation, such keyposes were chosen in the boundary of the robotic arm workspace, at two different height levels (8 per level). Figure~\ref{fig:poses_robot_collage} reveals the 16 proposed positions for the robotic arm, which are represented in the robot base coordinate system.



\begin{figure*}[ht]
\centering \includegraphics[width=\textwidth]{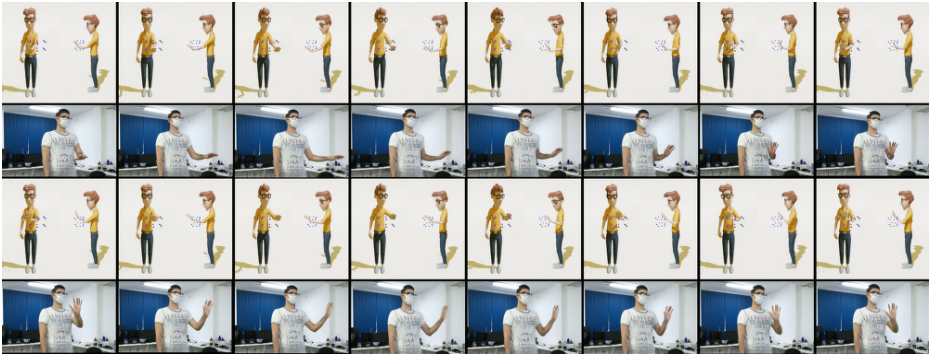}\\
\caption{All 16 poses used as reference for the TPS model training. For each avatar pose in the collage, a corresponding real user pose is shown below. Both avatar arm and user arm are mirrored. A set of 16 spheres with all avatar hand positions was used. Green sphere means current hand position, with blue spheres mean all remaining ones.}
\label{fig:poses_collage}
\end{figure*}

To assist the users in the training process, an interactive training tool is proposed. For each robotic arm keypose, an image is first rendered from a 3d human avatar performing the suggested corresponding human keyposes, as illustrated in Fig.~\ref{fig:poses_collage}. Then, the images are sequentially shown to guide the user in performing the pose. The user pose is collected using the depth sensor, mirrored (so that users think of the avatar as a mirror) and converted to the proposed arm representation (shown in Eq.~(\ref{eq:armlen_unormalized})). Since Kinect skeletons are prone to noise, a sequence of skeletons are collected over a specific period of time, and then the resulting skeleton is computed as the median skeleton, by taking the median value for each skeleton coordinate. 
It is worth mentioning that, differently from the naive affine transformation method, here the arm representation is not required to be normalized, since the TPS mapping is trained for each specific user body.

Finally, after the training process, the sets representing the robotic arm key poses ($Y$) and the collected user poses ($X$) are obtained to compute the TPS mapping, according to Eq.~(\ref{eq:tps_3}). In Section~\ref{sec:practical_experiments} we show that the resulting map is capable of performing smoothly while converting user arm poses to robotic arm poses.


\section{\textbf{Hand-state classification}}
\label{sec:classifier}
The second main part of our methodology aims to provide a natural way to control the robot gripper state using the user's hand, in a binary fashion (open or closed).

We focus on recognizing the user's hand state using the depth image. The advantage of this approach lies in its resilience against variations in color, texture, and lighting conditions. It is worth mentioning that, although hand skeletons can be estimated by SDKs such as the original Kinect V2 SDK, they are still very inaccurate due to the complexity of the hand skeleton and occlusion. Thus, since we aim to solve a binary classification problem, developing a binary classifier that exploits depth images was considered to be the most suitable approach, as already shown in \cite{visapp19}.

In that previous work, a  Convolutional Neural Network (CNN) was proposed to evaluate the user hand using a single depth image and map the corresponding state to the robot gripper in real-time. 
The main idea behind CNNs is that they learn to extract relevant features from input data, generating a proper feature map that allows the classification task to be optimally performed.
CNNs first perform convolutions on input data through several convolutional layers, possibly using max-pooling and dropout layers, and then flattening the resulting feature map into a single-dimension array. The flattened array is then classified using dense layers. 

Differently, in this work we propose to exploit a sequence of depth images collected from a short period, thus taking advantage of the temporal coherence of the problem. With this aim, we adopt a deep neural network architecture known as Long-term Recurrent Convolutional Network (LRCN, \cite{donahue2014}). In what follows, we first describe the depth image preprocessing, and then present the adopted LRCN architecture, highlighting its differences from the CNN presented in ~\cite{visapp19}.



\subsection{Depth image preprocessing}
\label{sec:preprocessing}
Depth images acquired by a depth sensor are first cropped, according to the hand position in the image space; filtered to remove background objects; and encoded in grayscale or binary image. This process is performed both in dataset acquisition (for classifier training) and during experiment tests.


The following data, which is provided by the Kinect v2 SDK in real-time, is exploited in this step:
\begin{itemize}
    \item raw depth image $d : U \subset \mathbb{R}^2 \mapsto \mathbb{R}$;
    \item 3d skeleton composed of 3d body joint positions in euclidian coordinates, where $j_i \in \mathbb{R}^3$ is the position of the $i$-th joint, in the Kinect camera space.
    \item 2d skeleton, computed by projecting the 3d skeleton joint positions onto the image plane, where $p_i \in U$ denotes the 2d coordinates, in pixels, of the $i$-th joint.
\end{itemize}

For the hand segmentation phase, we used the same approach as described in  \cite{visapp19}. Let $rh$ be the joint index of the user's right hand. The raw depth image $d$ is first cropped by defining a 2d squared hand region $\mathcal{H}$ centered at $p_{rh}$, with a side of $B$ pixels.
Note that, as the hand moves closer or further away, $\mathcal{H}$ must scale accordingly to keep the projected hand on a constant scale. More specifically, $B$ must vary inversely with the distance from the right hand to the sensor, which is given by $d(p_{rh})$.
Thus, we set the side of $\mathcal{H}$ according to the function $b : \mathbb{R} \mapsto \mathbb{R}$, which is given by
\begin{equation}
\label{eq:hand_bounding_box}
    b(d(p_{rh})) =\frac{K}{d(p_{rh})} \, ,
\end{equation}
where $K$ is an empirically tuned proportionality constant that we set to 60.

The raw depth image is cropped to the hand region $\mathcal{H}$, with side given by $b(d(p_{rh}))$, resulting in the cropped hand image $d' \colon \mathcal{R} \mapsto \mathbb{R}$. However, such an image may include unwanted data like other parts of the user body or background objects. To address this issue, two filtering approaches might be used: segmentation by depth thresholding; and segmentation followed by binarization. 


Depth thresholding is performed by simply computing a segmented image $g$ as follows:
\begin{equation}
\label{eq:segmentedimage}
    g(x,y) =
    \begin{cases}
        d'(x,y), & d'(x,y) \leq D_{min} + T \\
        0, & d'(x,y) > D_{min} + T
    \end{cases} \, ,
\end{equation}
where $T$ is a threshold whose value was empirically set to 0.2 meters; and $D_{min} = \min d'(x, y)$ is the closest point from the camera and, therefore, it is assumed to be a point of the hand. Thus, every point that is further than $D_{min} + T$ should be removed.

Optionally, a binarization procedure may be applied to generate a binary image $b$ according to:
\begin{equation}
\label{eq:binarizedimage}
    b(x,y) =
    \begin{cases}
        1, & d'(x,y) \leq D_{min} + T \\
        0, & d'(x,y) > D_{min} + T
    \end{cases} \, .
\end{equation}

In our work, we used the binarization approach. After generating $b(x, y)$, each image is resampled to $50\times 50$ pixels, which is the input format of the classifiers that we present and compare in this work. 

\subsection{Deep neural network architectures}
\label{sec:cnn_details}

\begin{figure*}[!t]
\centering \includegraphics[width=\textwidth]{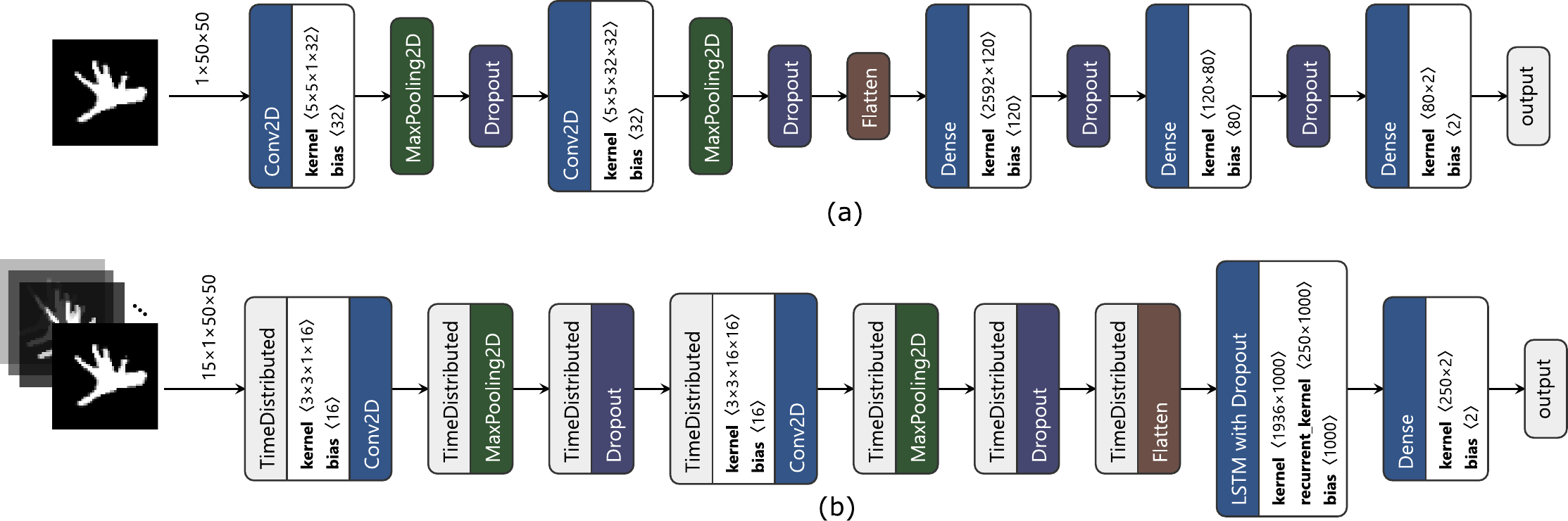}\\
\caption{Best CNN and LRCN classifiers found in the experiments described in Section~\ref{sec:results_classifier}. Images were generated with aid of Netron open-source software \cite{netronapp}.}
\label{fig:classifier_architectures}
\end{figure*}

In this subsection, we present the LRCN architecture that we adopt in the proposed methodology, and briefly describe the CNN presented in \cite{visapp19} which we use as a baseline to compare to the LRCN. Examples of each architecture are shown in Fig.~\ref{fig:classifier_architectures}.

It is worth mentioning that the CNN input is a single image, while the LRCN input is a sequence of many continuous images in time. More specifically, the LRCN classifier aims to predict the hand state of the last image of the input sequence.

\paragraph{Convolutional Neural Network.} In this approach, the preprocessed images described in Section \ref{sec:preprocessing} are given as input to a CNN, which is comprised of two parts: convolutional layers, to extract relevant features; and dense layers, to perform the binary classification. In the following, we describe hyperparameters that we experimented with in order to fairly compare CNNs to the proposed LRCN architecture.
The input data is filtered by 1 or 2 convolutional layers, where each layer may apply filters with size $(F)$ $3 \times 3$ or $5 \times 5$. For each convolutional layer, 8, 16, 32, 64, and 128 filters per layer $(S)$ were examined. After each convolutional layer, a max-pooling layer with $2\times2$ receptive fields may be applied to downsample the feature map. The resulting map is then flattened and follows to 1 or dense layers, with 80, 120, or 160 hidden units each $(D)$. Dropout probability $(P)$ defines both probabilities of dropout in convolutional and dense layers, in which case we tested the combinations $[0.5, 0], [0.5, 0.5]$ and $[0, 0.5]$. In all convolutional and dense layers, we use the rectified linear (\emph{ReLU}) as the activation function.





\paragraph{Long-term Recurrent Convolutional Network.} Long-term Recurrent Convolutional Network (LRCN, \cite{donahue2014}) may potentially exploit the temporal coherence in videos (or sequence of images). More specifically, since the hand state of the user continuously varies with time, we hypothesize that a classifier that exploits such coherence using a sequence of continuous images (LRCN), may achieve higher accuracy than a single-image classifier (CNN).
Different from the CNNs, the flattened feature map extracted from the convolutional layers are given as input to Long Short-Term Memory (LSTM) cells instead of dense units.
It is worth mentioning that, in order to train LRCN classifiers in a supervised manner, training set must comprise one binary label per image. Thus, after defining a fixed sequence length with value $n$, sequences of $n$ continuous images in time are compiled from the original training set, through a sliding window approach. Note that each sequence is associated with a single label, which must be the state of the hand at the last image of the sequence.
To perform a fair comparison, the convolutional layers hyperparameters were experimented as in the previous CNN architecture. In the recurrent part of these networks, a single LSTM-layer was experimented with a varying number of LSTM units: 50, 100, 250 and 500 units. Dropout probability $(P)$ was also experimented as in the CNN architecture. 
In all convolutional layers, we adopt the rectified linear (ReLU) as activation function and the hyperbolic tangent in all LSTM layers. Finally, a softmax dense layer with 2 neurons outputs the final probability distribution.

A grid-search was performed to evaluate variations of both CNN and LRCN architectures, to both tune and compare whether LRCN outperforms the CNN in the hand state recognition task. Results are revealed in Section~\ref{sec:results}.




\section{\textbf{System architecture}}
\label{sec:architecture}

This section describes implementation details of the proposed methodology. All modules were developed on a Windows 10, Intel Core i5, 16GB RAM DDR3, and NVidia GeForce GTX 2080 TI 12GB computer with ROS Melodic (Robot Operating System, \cite{ROS}). We used a robotic manipulator of model Denso VP6242 \cite{denso}, along with a Robotiq 2F-85 \cite{robotiq} gripper in our experiments.

The whole system was developed using two computers. One of them was responsible solely for direct communication with gripper and arm controllers through Matlab 2015. The other was responsible for running user data capture, classification and simulation modules.

To collect our dataset, we used a Microsoft Kinect V2 \cite{kinect} depth camera, including provided SDK (Standard Development Kit). The whole application was programmed in Python 3 language, except for the bridge between ROS and Kinect V2 SDK, which was programmed in C++. Our classification models were developed using TensorFlow GPU and Keras.

Modules from our system communicate in a distributed manner using the publisher/subscriber architecture provided by ROS, with TCP (Transmission Control Protocol) as the underlying communication protocol. In our study case, Windows was chosen because Kinect SDK V2 is unavailable for Linux. There are, however, viable alternatives such as Libfreenect2 from OpenKinect community \cite{xiang2016libfreenect2} and OpenNI \cite{villaroman2011teaching} libraries for Kinect data manipulation.

How the system was developed allowed for decoupling system components. Classification modules are the same for both visual simulation and real-life trials. The complete application comprises the following modules:
\begin{itemize}
    \item Kinect V2 and ROS interface;
    \item Database acquisition;
    \item Hand-state classification model;
    \item Hand-position transformation model;
    \item Robot visual simulation;
    \item Real robot control;
\end{itemize}

Hand-state classification and hand-position transformation models can be set in run-time, via ROS parameters. Simulation and robot control nodes can be turned on and off without interfering in classification, since both nodes behave as read-only consumers from ROS topics. A detailed flow diagram of the system is shown in Fig. \ref{fig:diagram}. Data is passed from left to right, except when persisting input data is desired. The complete system is capable of running in real-time, at 30 frames per second.



\section{Experiments}
\label{sec:results}

In this section, we describe the experiments performed to both validate and evaluate the proposed methodology and discuss the results achieved. We first describe experiments to compare the novel LRCN hand state classifier and the previously presented CNN (Section~\ref{sec:results_classifier}). Then, we describe the setup of user experiments performed to evaluate and compare the proposed pose mapping approach in Section~\ref{sec:practical_experiments}. Finally, results of the latter experiments are revealed and discussed in Section~\ref{sec:results_mapping}.

\subsection{Hand State Classifier}
\label{sec:results_classifier}

We collected a database of binary images comprised of 9668 labeled images from 7 different people and organized them into 14 sequential episodes, publicly available~\footnote{\url{https://drive.google.com/file/d/1bOQfOFtDxgGfp_b9HELAK3WNchBUxGdX}}. As the same database would serve both for the CNN and the LRCN methods, it was organized in an episodic way. For each person, two episodes were continuously recorded at 30 FPS while the user performed the grab-and-release task (imagining an object in the air) from one side to another. The difference between each of the two episodes per person was the starting side for the grab-and-release movement.

After collecting the database, the next step was to train the classifiers. We adopted a 2-fold cross-validation setting to split our available data, in which one fold was used for training and the other was used for testing. A grid search was conducted over the hyperparameters previously described for both the CNNs and the LRCNs (Section~\ref{sec:cnn_details}). In this section, we present the best configurations found for both variants.

CNN performed better when using dropout probability $P$ of 0.5 in both convolutional and hidden layers. Table \ref{tab:best_cnn_accuracies} reveal the results for the experimented CNNs. The best CNN architecture showed to be for convolutional kernel sizes $F = [[5, 5], [5, 5]]$, with $S = 32$ filters per layer, and two hidden layers of dimensions $D = [120, 80]$. It is also worth mentioning the similar accuracy achieved by the top-performing architectures, suggesting low sensitiveness to hyperparameter variations over the best classified ones.


\begin{table}[!b]
\footnotesize
\caption{ Results for the top 10 CNN-based networks in our grid-search, sorted by accuracy. }
\label{tab:best_cnn_accuracies}
\begin{center}
\scalebox{1}{
\begin{tabular}{c c c c | c | c | c }
$F$ & $S$ & $D$ & $P$ & weights & accuracy & loss\\
\hline
[[5, 5], [5, 5]] & 32 &     [120, 80] & [0.5, 0.5] &    347,466  & 96.5349 & 0.1294 \\

[[5, 5], [5, 5]] &	64 &	[160, 160] & [0.5, 0.5] &	959,810	 &  96.5246 &	0.1308 \\

[[5, 5], [5, 5]] &	16 &	[160, 80]  & [0.5, 0.5] &	227,394  &	96.5246 &	0.1210 \\

[[5, 5], [5, 5]] &	64 &	[120, 80]  & [0.5, 0.5] &	736,170  &	96.5142 &	0.1280 \\

[[5, 5], [5, 5]] &	32 &	[160, 80]  & [0.5, 0.5] &	454,386  &	96.5142 &	0.1241 \\

[[5, 5], [3, 3]] &	64 &	[160, 0]   & [0.5, 0.5] &	1,063,074 &	96.4418 &	0.1327 \\

[[5, 5], [3, 3]] &	64 &	[160, 80]  & [0.5, 0.5] &	1,075,794 &	96.4315 &	0.1350 \\

[[5, 5], [3, 3]] &	16 &	[160, 80]  & [0.5, 0.5] &	271,938	  & 96.4315 &	0.1136 \\

[[5, 5], [3, 3]] &	64 &	[160, 80]  & [0, 0]     &	1,075,794 & 96.4108 &   0.3058 \\

[[5, 5], [5, 5]] &	128 & 	[160, 80]  & [0.5, 0.5] &	2,085,138 &	96.4004 &	0.1502 \\

\hline
\\
\end{tabular}
}
\end{center}
\end{table}

\begin{table}[!b]
\footnotesize
\caption{ Results for the top 10 LRCN-based netweorks in our grid-search, sorted by accuracy. }
\label{tab:best_lstm_accuracies}
\begin{center}
\scalebox{1}{
\begin{tabular}{c c c c | c | c | c }
$F$ & $S$ & $D$ & $P$ & weights & accuracy & loss\\
\hline
[[3, 3], [3, 3]] &	16 &	250 &	[0.5, 0.5]  &	2,189,982 &	98.7225 &	0.0494 \\

[[3, 3], [5, 5]] &	128 &	50 &	[0, 0]      &	2,981,310 &	98.7225 &	0.0413 \\

[[5, 5], [3, 3]] &	64 &	250 &	[0.5, 0.5]  &	6,690,094 &	98.7119 &	0.0541 \\

[[3, 3], [5, 5]] &	32 &	250 &	[0.5, 0.5]  &	3,477,454 &	98.7119 &	0.0503 \\

[[3, 3], [5, 5]] &	32 &	50 &	[0, 0]      &	676,254 &	98.6908 &	0.0438 \\

[[5, 5], [5, 5]] &	128 &	250 &	[0, 0]      &   11,032,558 & 98.6908 &	0.0428 \\

[[5, 5], [3, 3]] &	32 &	100 &	[0.5, 0.5]  &	1,330,682 &	98.6908 &	0.0497 \\

[[3, 3], [5, 5]] &	128 &	100 &	[0, 0]      &	5,571,610 &	98.6908 &	0.0433 \\

[[3, 3], [5, 5]] &	16 &	500 &	[0.5, 0.5]  &	4,209,578 &	98.6803 &	0.0564 \\

[[5, 5], [3, 3]] &	64 &	50 &	[0.5, 0.5]  &	1,328,894 &	98.6803 &	0.0461 \\

\hline
\\
\end{tabular}
}
\end{center}
\end{table}

A similar grid search was performed to find the best LRCNs classifiers. Results are shown in Table~\ref{tab:best_lstm_accuracies}. Dropout showed a weaker influence on LRCN accuracy. Two specific configurations achieved the same top accuracy value of 98.72\% and thus outperforming the best CNNs. This is one of the most relevant results of our experiments, suggesting that, by exploiting a sequence of many images, LRCNs are capable of coping with challenging examples.

To better investigate the effects of the sequence length, we experimented with the top-performing LRCN architecture on three distinct sequence lengths. The results are shown in Table~\ref{tab:lstm_seq_len}, revealing higher accuracy for longer sequences. Thus, we conclude that accuracy is improved when more data is available to be evaluated by the LRCNs.

\begin{table}[!t]
\caption{Configuration and accuracy of the best-performing LRCN architecture and the sequence length.} 
\label{tab:lstm_seq_len}
\begin{center}
\begin{tabular}{c|c|c}
sequence len. & accuracy & loss   \\ \hline
15            & 98.7225  & 0.0494 \\
10            & 98.3337  & 0.0575 \\
5             & 98.1377  & 0.0699  \\
\hline
\end{tabular}
\end{center}
\end{table}

\subsection{User Experiments Setup}
\label{sec:practical_experiments}

To validate and compare the proposed Thin Plate Splines Pose Mapping (TPS-PM, Sections~\ref{sec:tps} and \ref{sec:customtps}) to the previously proposed Naive Affine Pose Mapping (NA-PM, Section~\ref{sec:naivemapping}), we performed usability tests to measure the user experience with both compared approaches. We set up a test environment and predetermined a set of five tasks in which users were asked to manipulate objects with the robotic arm through both pose mapping approaches.

For each position mapping approach, the following aspects were evaluated:
\begin{itemize}
    \item \textbf{Training}. Were users able to accomplish the training procedure successfully?
    \item \textbf{Goal achievement}. Were users able to accomplish the required tasks?
    \item \textbf{Comfort}. Which approach feels more comfortable and natural?
    \item \textbf{Learning curve}. Does the learning curve allow rapid progress in such a way that users could cope with increasingly difficult tasks?
\end{itemize}

\begin{figure}[ht]
\centering \includegraphics[width=0.8\linewidth]{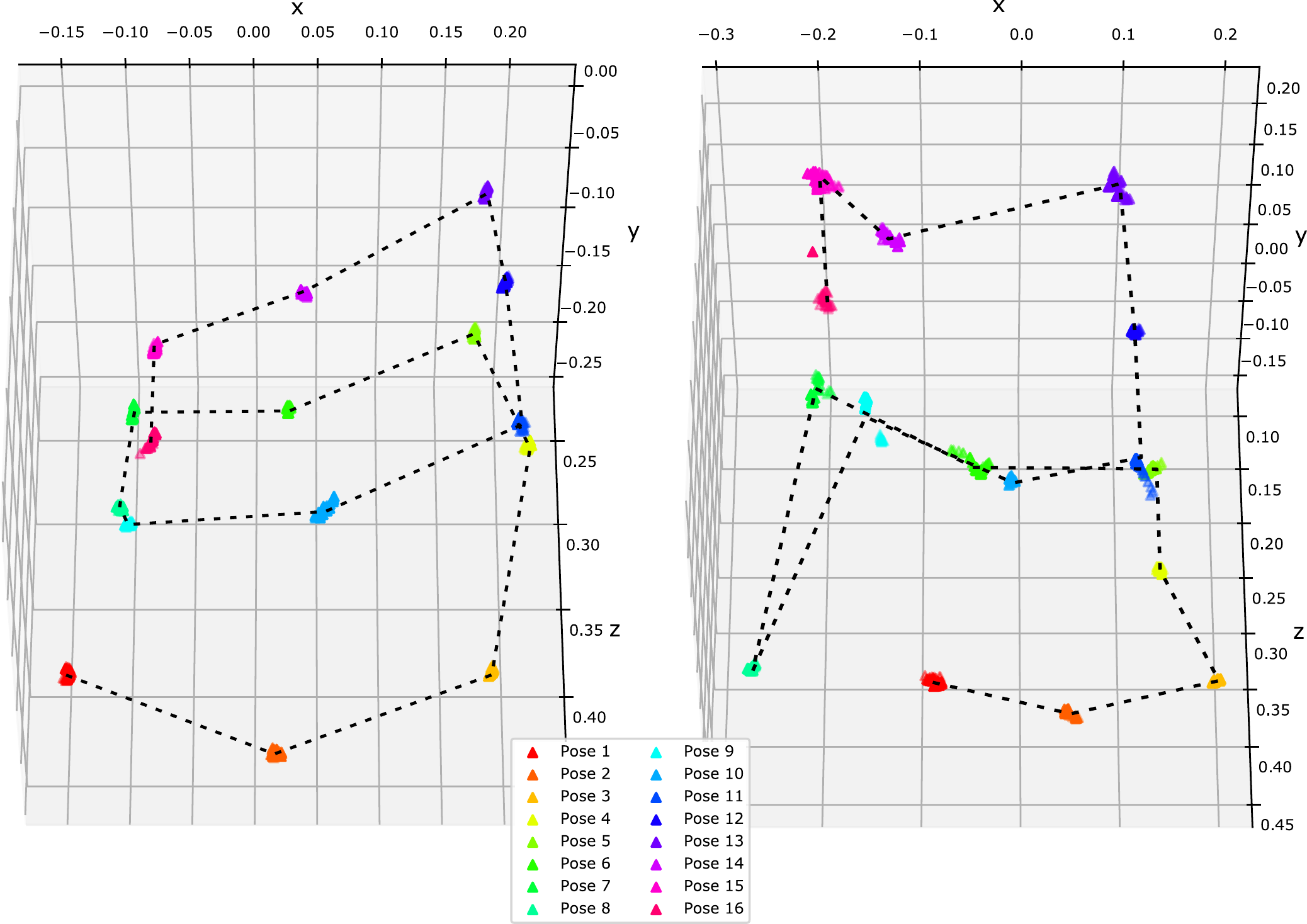}
\caption{Three-dimensional visualization of a successful (left) and of a rejected  (right) TPS-PM customization. The difference is noticeable in the relative positions between consecutive poses. A low-quality calibration will not follow the relative placement from one pose to another, generating an ill formed model.}
\label{fig:calibration_samples}
\end{figure}

At the beginning of each session, users were asked to perform the training to calibrate their body to each pose mapping approach (NA-PM and TPS-PM). For the NA-PM, the calibration was effortless, since the only requirement from users was to stretch their right arm in order to measure its length. In contrast, TPS-PM calibration was more time-consuming. The main issue was that TPS-PM required a considerable spatial awareness from users to correctly perform the keyposes during the training process. Thus, it was necessary to reject low-quality TPS-PM customizations. More specifically, a supervisor visually examined the sixteen collected hand positions to check whether they had an unusual spatial distribution, with two or more vertices being too close together or in an unexpected relative position. 
Figure \ref{fig:calibration_samples} shows an example of a successful and of a failed TPS-PM customization, respectively. When the training failed, the user was asked to repeat the procedure.

The set of experimental tasks required not only motor skills and spatial awareness from the users but also the ability to adapt to the specific transformation being used. Each experiment has some increased difficulty in relation to the previous one. Before the experiments, users were asked to fill out a compliance form allowing them to be recorded during the sections for analysis and scientific purposes.

Interaction elements employed in the experiments were:
\begin{itemize}
    \item two cubes of length equal to 5 cm;
    \item a cylindrical platform with a red-marked open-bottom of 10 cm diameter and a height of 12 cm. The top has 7.5 cm of diameter.
\end{itemize}
When the platform was placed upside-down, it became a hoop. We also defined 4 areas within the robot workspace, named from A to D.

To quantify each task, we use the concept of atomic movements. One atomic movement is an indivisible continuous displacement of the robot gripper with well-defined starting and ending times. Translations and gripper state variations (opening/closing, or grab/release) are also considered atomic movements. Thus, each task can be performed using a minimum number of atomic movements.

More specifically, tasks were:

\begin{enumerate}
    \item \textbf{Take a cube from region B to region A}. It's an initial task to test the user's ability to successfully grab and move an object to another location. Can be achieved with a minimum of 4 atomic movements: home-to-B, grab, B-to-A, release.
    \item \textbf{Swap cubes between regions B and A, using region C as auxiliary}. The purpose of this task is to measure the increase in user's skills since the first task. The swap task can be done in a minimum of 12 atomic movements, all of them with the same difficulty as the first task, e.g.: home-to-A, grab, A-to-C, release, C-to-B, grab, B-to-A, release, A-to-C, grab, C-to-B, release.
    \item \textbf{Bring cubes from regions A and B to the hoop}. This task introduces the necessity of motion at higher heights, thus increasing task complexity in the vertical axis. Users have to take into consideration avoiding collision between the gripper and the hoop while bringing the cubes to the proper height before releasing it. Skills evaluated here also include ease of use near robot boundaries, since the platform is located at region D. It can also be executed with a minimum of 8 atomic movements, e.g: home-to-A, grab, A-to-hoop, release, hoop-to-B, grab, B-to-hoop, release.
    \item \textbf{Pile cubes in region A}. This task introduces a new component in evaluation which is the ability to fine-tune the cube position to successfully pile one cube onto another. Previous tasks had a less strict acceptance with relation to where the cubes might be placed. Can also be performed within 8 atomic movements, e.g.: home-to-C, grab, C-to-A, release, A-to-B, grab, B-to-A, release.
    \item \textbf{Move a cube from platform to A, then from B to platform.} Skills needed are boundary regions control and the ability to fine-tune the cube position to place it on top of the platform. Can also be performed on a minimum of 8 atomic movements, e.g.: home-to-hoop, grab, hoop-to-A, release, A-to-B, grab, B-to-hoop, release.
\end{enumerate}

\begin{figure}[h]
\centering \includegraphics[width=0.6\linewidth]{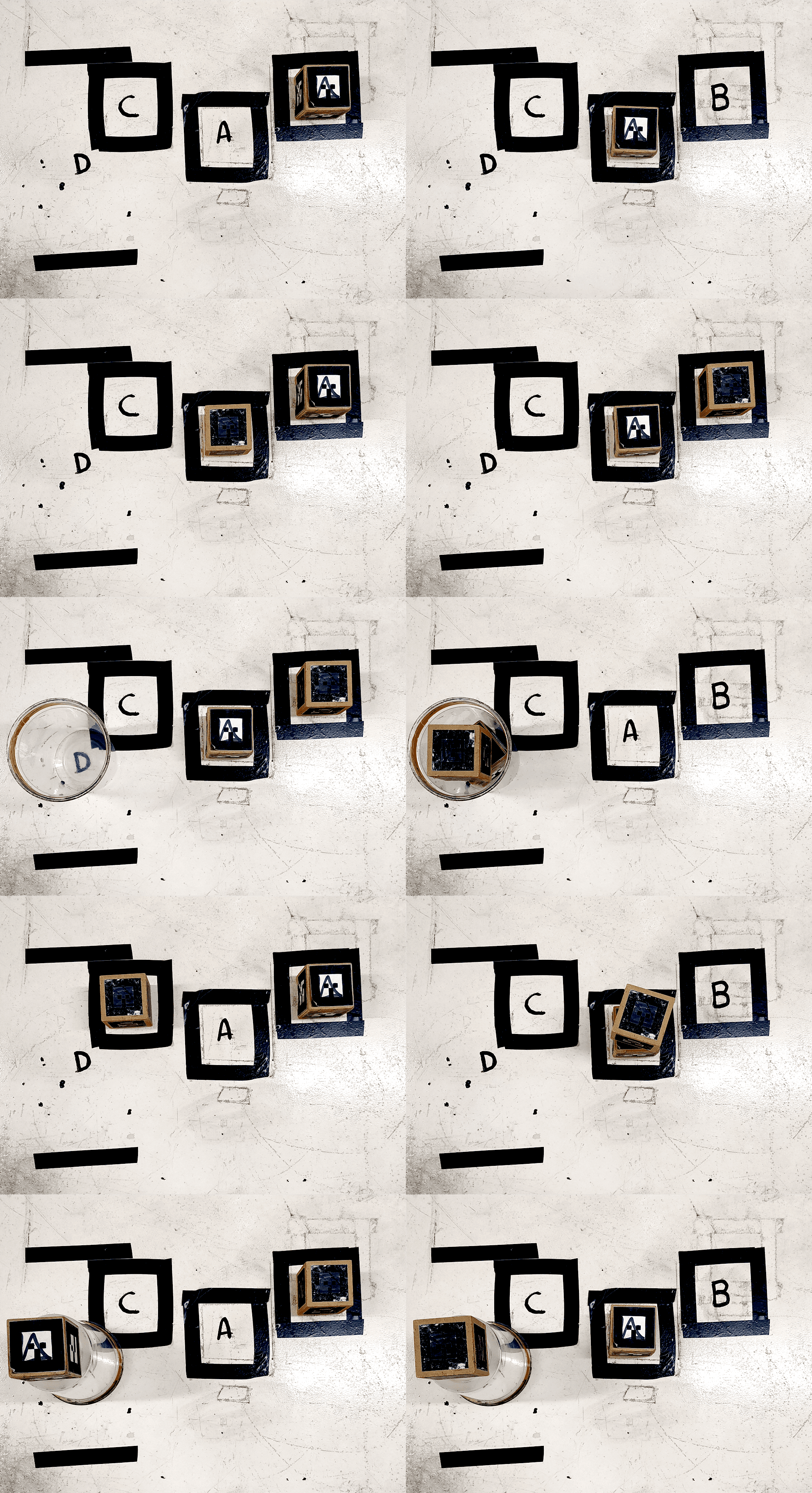}\\
\caption{Initial (left) and target (right) configurations of each task performed by the users. Each row represents one task from 1 to 5, respectively.}
\label{fig:experiments}
\end{figure}

Both initial and target configurations for each of the previously discussed experiments are shown in Fig.~\ref{fig:experiments}. In order to measure time completion for each experiment, we defined the following conditions: (1) time counter starts as soon as the user-control mode is triggered, and; (2) time counter finishes as soon as the target is reached and the objects of interaction remain still. Success occurs when the target configuration is reached and objects stay within established areas (not touching the workspace region outside the black markers). A fail occurs when the experiment reaches a state where the target configuration is not feasible without human assistance (\emph{i.e.} when one cube falls outside the robot workspace). 

The previously discussed set of five experiments was repeated twice for each user, firstly using NA-PM and then using TPS-PM. Before the very first experiment of each pose mapping method started, a ``playground time'' of two minutes was given to each user for free operation so that they could acquire familiarity with the method. After the very last experiment of each session, users were asked to answer two multiple choices quantitative Likert-based \cite{likert1932technique} questions by selecting a value in the range from 1 to 5, about how they felt with relation to the pose mapping method:
\begin{itemize}
    \item \textbf{(Q1) How smooth was the movement of the robotic arm during control?} This question evaluates how comfortable users were with the smoothness of the robotic arm movement when his right hand moves. 
    \item \textbf{(Q2) How difficult was the movement near the workspace boundaries?} This question evaluates the quality of the pose mapping when the robot gripper is closer to its boundaries. A good mapping would present the user with a consistent low difficulty that does not depend on the region of operation, with an overall effort being as minimum as possible.
\end{itemize}

After both sets of experiments were finished and their respective objective questions were answered, the users were asked to answer a set of 3 discursive questions to evaluate overall opinion about the comparison between both mappings:

\begin{itemize}
    \item \textbf{(Q3) In your opinion, which were the advantages of each mapping?}
    \item \textbf{(Q4) What were the challenges found during the experiment?}
    \item \textbf{(Q5) Would you have any suggestions to improve the interface?}
\end{itemize}

All participant answers were collected through an online form for posterior analysis. At this phase, five participants volunteered for testing purposes. A two-sample Student's T-test~\cite{student1908probable} was performed over collected duration/atomic values for the two classes (NA-PM and TPS-PM).

\subsection{User Experiments Results}
\label{sec:results_mapping}

After successfully performing the customized training procedure, each individual executed each of the five proposed tasks using NA-PM, before executing the same tasks using TPS-PM. Table~\ref{tab:experiment_lenghts} shows the time duration taken by each participant to complete each task. We did not ask participants to perform the tasks as fast as possible, thus allowing them to take the necessary time without a rush. All proposed tasks were able to be completed within 3 or 
fewer trials. A sample of each task may be seen in a public available video~\footnote{Video with experiments: \url{https://youtu.be/Rk3iS_KnaWc}}.

\begin{table*}[ht]
\caption{Timing results for each experiment performed by the volunteers. ``F'' means fail in the current trial, with each specific task being offered a maximum of 3 trials. Times are represented in seconds from the beginning to the completion.}
\label{tab:experiment_lenghts}
\small\addtolength{\tabcolsep}{-2.5pt}
\begin{tabular}{cccccc|ccccc}
\cline{2-11}
& \multicolumn{5}{c|}{duration (secs) for NA-PM} & \multicolumn{5}{c}{duration (secs) for TPS-PM}      \\ \hline
\multicolumn{1}{c|}{part. time/task} & 1      & 2      & 3         & 4     & 5     & 1     & 2        & 3     & 4           & 5        \\ \hline
\multicolumn{1}{c|}{p1}                     & 13.3  & 29.2  & 29.1     & 23.0 & 38.2 & 11.0 & 43.1    & 25.2 & 21.2       & F, 26.2 \\
\multicolumn{1}{c|}{p2}                     & 14.1  & 33.1  & F, 30.0  & 34.1 & 25.1 & 11.2 & 32.1    & 36.0 & 22.1       & 28.2    \\
\multicolumn{1}{c|}{p3}                     & 12.2  & 43.1  & F, 36.1  & 23.2 & 32.2 & 19.1 & 76.0    & 51.1 & F, F, 31.0 & F, 35.0 \\
\multicolumn{1}{c|}{p4}                     & 20.1  & 47.2  & 28.1     & 34.2 & 28.2 & 17.3 & 54.0    & 29.1 & 59.3       & 35.0    \\
\multicolumn{1}{c|}{p5}                     & 08.1  & 21.0  & 15.1     & 13.2 & 20.1 & 09.0 & F, 15.0 & 23.0 & 16.2       & 22.2    \\ \hline
\multicolumn{1}{c|}{mean}                     & 13.3 &	33.1 &	29.1 &	23.2 &	28.2 &	11.2 &	43.1 &	29.1 &	22.1 &	28.2   \\
\multicolumn{1}{c|}{std}                     & 4.3 &	10.6 &	7.7 &	8.8 &	6.9 &	4.4 &	23.0 &	11.3 &	17.2 &	5.6   \\ \hline
\multicolumn{1}{c|}{atm}                & 4     &	12    &	8    &	    8 &	    8 & 	4 & 	12 &    8 &     	8 &	8   \\
\multicolumn{1}{c|}{mean/atm}            & 3.3  &   2.8  & 3.6 &	2.9 &	3.5 &	2.8 &	3.6 &	3.6 &	2.8 &	3.5 \\ \hline

\end{tabular}
\end{table*}

\begin{figure}[ht]
\centering \includegraphics[width=0.9\linewidth]{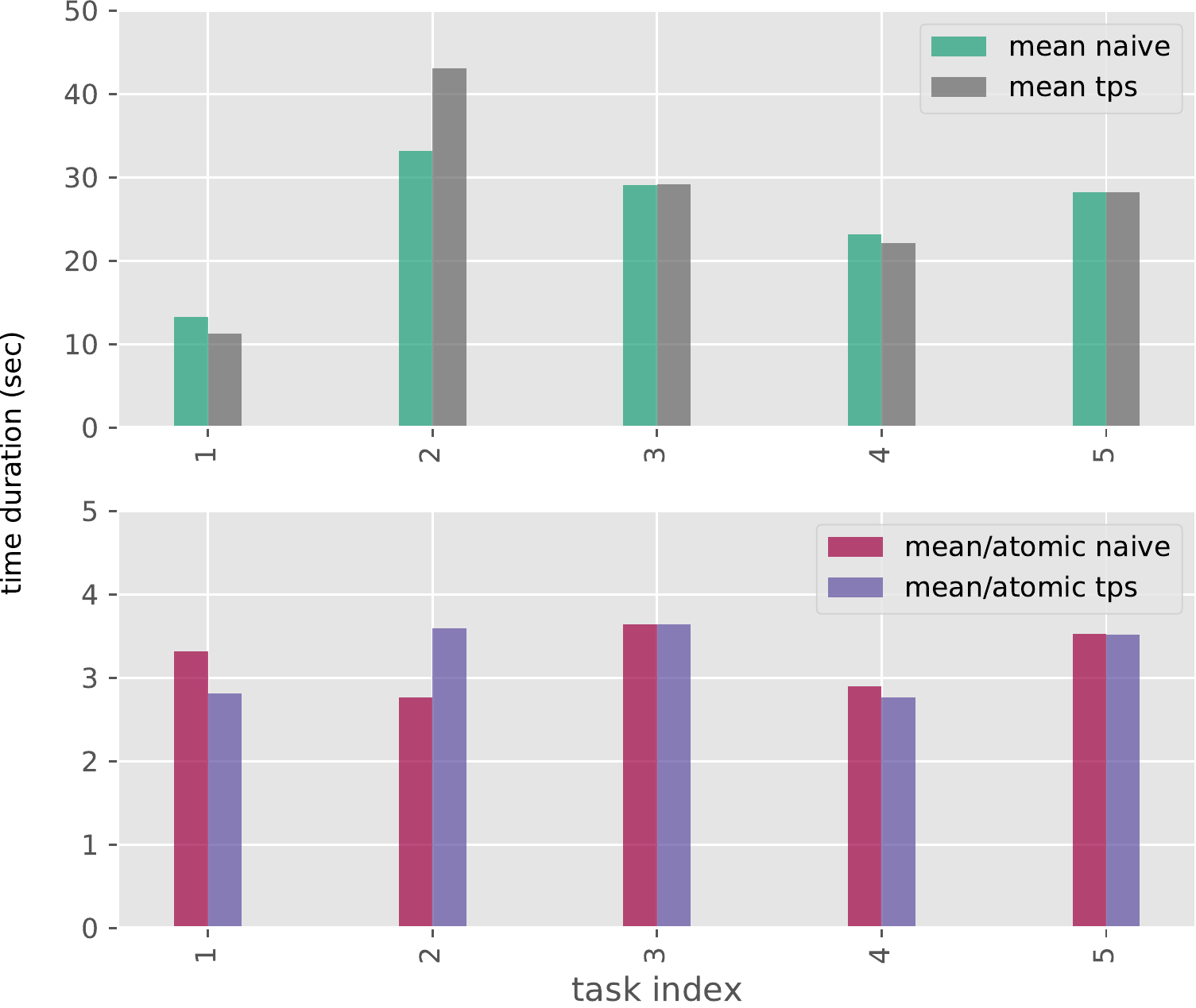}\\
\caption{Comparison between durations for both control modes (NA-PM and TPS-PM). Showing both the mean duration and the mean duration per atomic movement values. Graphs does not show substantial alterations in duration values, which leads to believe that novel method does not increase task difficulty.}
\label{fig:time_durations_barplot}
\end{figure}

\begin{figure}[ht]
\centering \includegraphics[width=0.9\linewidth]{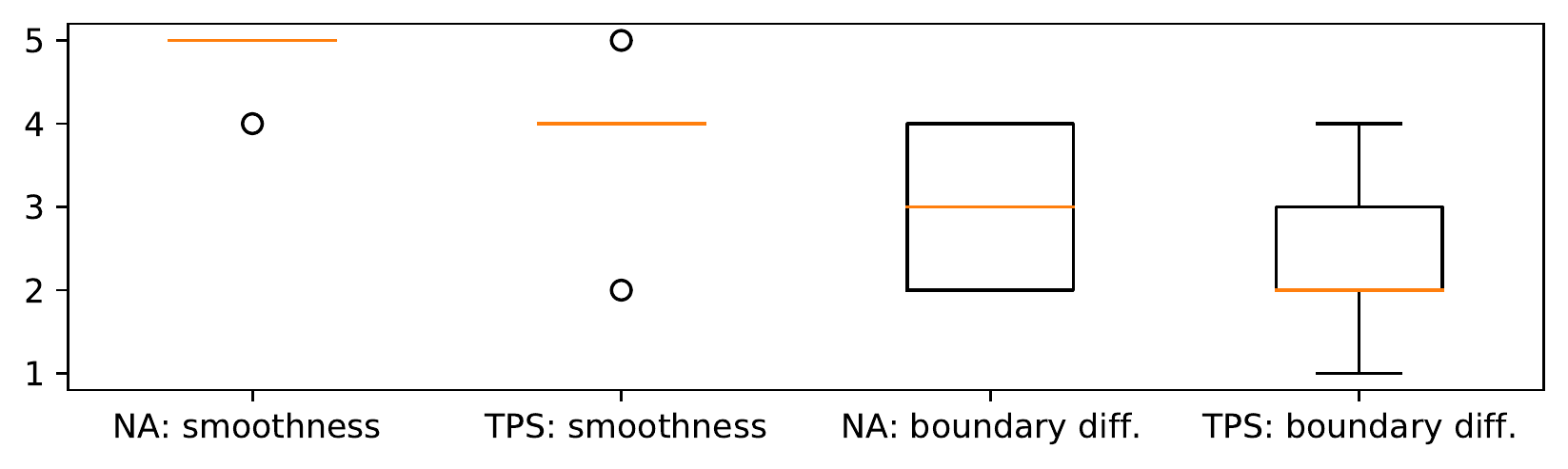}\\
\caption{Comparison between both control modes (NA-PM and TPS-PM). Both questions Q1 and Q2 were answered in a scale of intensity from 1 to 5. Naive Affine mode performance was better with respect to smoothness, according to users. Thin Plate Spline performance with respect to difficulty was better near to boundary limits of robot arm workspace.}
\label{fig:boxplot_smooth_diff}
\end{figure}

By analysing the time intervals of Table \ref{tab:experiment_lenghts}, it's possible to infer that the change on the control mode did not have a substantial influence over the total duration of each task. To further investigate this fact, we conducted a two-sample T-test over both classes, for the mean/atomics variable, assuming that classes were independent. The T-test resulted in a p-value of $0.899 > 0.05$, which confirms the null hypothesis in which both mapping modes (NA-PM and TPS-PM) have the same expectation for duration/atomics values. Such fact leads to the assumption that both modes have a similar learning curve, even though each one of them presents its own peculiarities. Such fact can be confirmed visually as shown in Figure \ref{fig:time_durations_barplot}.

Some of the participants were able to perform the tasks faster, showing a better spatial awareness, as the participant p5, for instance. Others presented a more cautious and curious manner, trying to experiment with the nuances of the robot movement, as the participant p3, for instance. This heterogeneity on the test participants contributed to a substantial diversity in the final duration of the tasks.

Tasks 1, 2, and 4, which did not require operation in high heights, nor interaction with the hopper/platform, presented the best duration per atomic operation. Tasks 1 and 2 were able to be successfully fulfilled within one single trial for all participants, except one. Tasks 3 and 5 failures were due to the collision of the robot gripper with the hopper/platform, causing a displacement since it was not fixed to the environment, leading to an unrecoverable state of the current task. Failures in Task 4 were due to a specific displacement in the center of the user's hand position when the hand state was toggled, \emph{i.e.} when the hand closes and its center moves closer to the user's wrist. User p3, with his cautious explorer profile, tried to understand this behavior during the experiment, causing him to take longer to complete the task. 

Participants' answers for the subjective questions Q1 and Q2 are presented in Figure \ref{fig:boxplot_smooth_diff}. From these plots only, it is suggested that participants agreed more with respect to smoothness of the robot arm movement. On the other hand, they presented more variance within answers related to the difficulty of arm control near the boundaries of the workspace, but suggested a tendency for a decrease in difficulty.

In general, users considered the first control mode to be smoother concerning the robotic arm speed, but found the second control mode to be better near the robot workspace boundaries. Both phenomena are explained by the fact that TPS-PM allows users to define their custom mapping in a non-linear fashion. Thus, while this customization provides a more accurate mapping in the boundary of the workspace, its non-linearity also implies variations in the magnitude of the gradient of the mapping function, which harms the sense of smoothness. Also, it was noted that smaller users' workspace volumes resulted in more comfortable control of the robot since users do not need to completely stretch their arm to reach robot arm boundaries. In other words, users only need to reach the boundaries of their trained workspace model.

As questions Q3 to Q5 were open and subjective, we will discuss the most frequent statements from the participants.
Question Q3 was related to the advantages of each mapping. According to the users, the advantages of the NA-PM model were intuitiveness and smoothness. One participant reported that ``\textit{The advantage of NA-PM is the displacement of the arm matching the displacement of the robot, in other words, the relative movement is almost the same. The advantage of the TPS-PM is the fact of the movement at boundaries did not degenerate}''.

Question Q4 was related to the challenges faced during task executions, where answers presented the higher diversity among the three subjective questions. One participant stated difficulties related to the hand-state classifier. We concluded that this difficulty was due to the hand bounding-box size (defined by Eq.~(\ref{eq:hand_bounding_box}) not taking into consideration users' hand size, but only its distance from the camera. Thus, users with excessive small or bigger hands, which were not present in the hand state classifier training dataset, would face a decreased accuracy in the classifier. This issue can be attenuated by training custom bounding box sizes for each user.

Another challenge for the users was to learn both mappings variants in consecutive sessions. This might have lead to confusion and thus less-than-optimal performance.

Finally, question Q5 was related to improvement suggestions. One suggestion was to have a classifier that would work with any hand orientation. In current work, for the hand-state classifier to be accurate, participants were asked to maintain their hands at a specific orientation (faced to the depth camera). Another valid suggestion was to create a more intuitive calibration for the TPS model, with some kind of online visual feedback, leading to an easier calibration process. 

Overall, we conclude that, while the NA-PM is efficient and practical for many simple tasks, the novel TPS-PM approach is more appropriate for tasks where high precision near the workspace boundaries is required. Besides, TPS-PM still provides reliable performance for straightforward tasks.

\section{Conclusion}
\label{sec:conclusion}

In this work we presented a novel approach for kinematic mapping between human and robotic arm poses that combines a  user-customizable pose mapping method based on Thin Plate Splines (TPS); and Long-term Recurrent Convolutional Networks (LRCNs) for hand state recognition. A complete motion-based teleoperation system using only a depth camera was built and validated through user experiments. The system works in a distributed way, where decoupled modules communicated through Robot Operating System (ROS) framework. Cross-validation experiments showed higher accuracy of the LRCNs networks when compared to a previously proposed CNN classifier to recognize human pose state. Besides, user experiments revealed that users adapted successfully to the new approach. Although the timings to accomplish the proposed tasks were similar to the compared straightforward affine mapping, it was perceived by the users that the TPS pose mapping was more accurate in regions closer to the workspace boundaries.

\section*{Acknowledgements}
\label{sec:acknowledgements}

The authors would like to thank: (1) FAPEAL/CAPES grant 05/2018 for funding and supporting the research; (2) The Instituto de Computação (IC/UFAL) for providing the necessary infrastructure for the development of this project.

%
%


\bibliographystyle{spmpsci}      

\bibliography{main.bib}   


\end{document}